\documentclass{article} 
\usepackage{iclr2027_conference,times}

\usepackage{amsmath,amsfonts,bm}

\def\eqref#1{equation~\ref{#1}}

\def\1{\bm{1}}

\DeclareMathAlphabet{\mathsfit}{\encodingdefault}{\sfdefault}{m}{sl}
\SetMathAlphabet{\mathsfit}{bold}{\encodingdefault}{\sfdefault}{bx}{n}

\usepackage{hyperref}
\usepackage{url}
\usepackage[dvipsnames]{xcolor}
\usepackage{booktabs}
\usepackage{multirow}
\usepackage{graphicx}
\usepackage{subcaption}
\usepackage{wrapfig}
\usepackage{capt-of}
\usepackage{amsthm}
\usepackage{amsthm}
\newtheorem{lemma}{Lemma}
\newtheorem{corollary}{Corollary}
\usepackage{caption}
\usepackage{makecell}

\title{Attend at the Right Resolution: Head-Adaptive Sparse Attention via Query–Key Interaction Structure}
\author{
Wajih Hassan Raza$^{1}$, Ahsan Bilal$^{2}$, Yidan Shen$^{1}$, Renjie Hu$^{1}$, Xin Fu$^{1}$, Yu Wen$^{1}$ \\
$^{1}$University of Houston \quad $^{2}$University of Oklahoma
}

\iclrfinalcopy 
\begin{document}

\maketitle
\lhead{Preprint}

\begin{abstract}

Attention is a major scalability bottleneck in large language models (LLMs), with its computational cost and key-value (KV) memory traffic becoming increasingly prohibitive as context length grows. Sparse attention addresses this bottleneck by restricting each query to a subset of the context, but its effectiveness hinges on identifying relevant regions both cheaply and reliably. Existing training-free methods primarily rely on runtime relevance estimates, with selection dimensionality and granularity typically governed by fixed heuristics rather than the intrinsic interaction structure of pretrained attention heads. We take a different perspective: the query–key operator of each pretrained head defines a characteristic response structure that determines the resolution needed for sparse selection, both across feature directions and across context tokens. Building on this insight, we propose Head-Adaptive Sparse Attention (HASA), a training-free sparse attention framework that jointly adapts selection resolution along these two dimensions. Along the feature dimension, the singular spectrum of each head's query–key operator determines an effective rank, yielding a compact, head-specific response space. Along the sequence dimension, neighboring regions are progressively merged according to the directional coherence of their token responses in this same space, while heterogeneous regions remain separate. Exact attention is then computed using the original full-dimensional KV states of the selected regions. HASA captures \textbf{88.1\%} of dense-attention mass and, compared with UNIQUE, a strong recent sparse-attention baseline, improves accuracy by up to 3.47 and 2.57 points on RULER and LongBench-Pro, respectively, and accelerates self-attention by up to \textbf{1.32$\times$}, while achieving up to \textbf{14.07$\times$} speedup over dense attention.

\end{abstract}



\section{Introduction}
Large language models (LLMs) have become a general-purpose foundation for a wide range of applications, with attention serving as the core mechanism for integrating information across the input context. However, dense attention requires each query to interact with the entire available context, incurring substantial computational cost and key-value (KV) memory traffic as sequence lengths and inference workloads increase \citep{achiam2023gpt4,dubey2024llama3,comanici2025gemini,pope2023scaling,dao2024flashattention2,kwon2023pagedattention}. These growing costs make efficient attention increasingly important for scaling LLM inference to longer contexts and larger workloads.

\begin{wrapfigure}{r}{0.50\columnwidth}
    \vspace{-15pt}
    \centering
    \includegraphics[width=\linewidth]{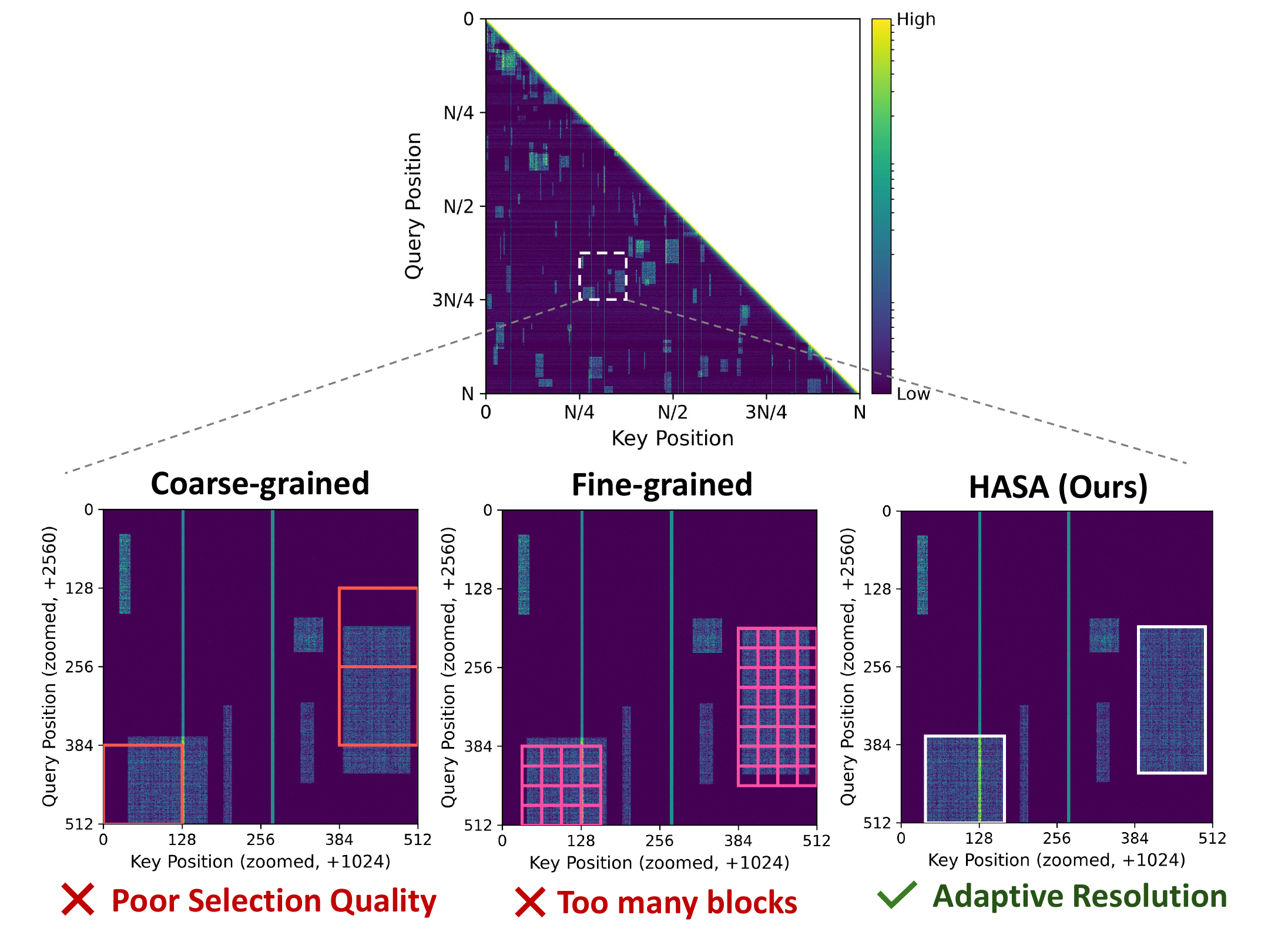}
    \caption{\textbf{Motivation for HASA.} Coarse-grained selection risks missing relevant tokens, while fine-grained selection increases selection overhead. HASA adapts selection granularity for each head.}
    \label{fig:motivation}
    \vspace{-10pt}
\end{wrapfigure}

Sparse attention reduces these costs by exploiting the observation that attention is often highly concentrated, allowing computation and KV access to be restricted to a small subset of the context while preserving most relevant interactions. Existing methods broadly follow two strategies. Approaches without per-query selection impose predefined access patterns or reduce the retained KV context using structural or historical signals, offering regular and predictable computation but limited ability to adapt to the interactions required by each query \citep{beltagy2020longformer,xiao2024streamingllm,xiao2024duoattention,zhang2023h2o,li2024snapkv}. Content-adaptive approaches, in contrast, estimate query-dependent relevance from runtime information to select relevant tokens, enabling sparse patterns to better capture input-dependent and long-range dependencies, albeit at the cost of additional selection computation \citep{yuan2025nsa,zhao2025infllmv2}. Despite these differences, both strategies face a fundamental challenge in determining the appropriate resolution for context selection. As illustrated in Fig.~\ref{fig:motivation}, coarse-grained selection groups relevant and irrelevant tokens into the same region, forcing a choice between accessing unnecessary KV states and missing important interactions. Fine-grained selection localizes relevant tokens more precisely, but requires evaluating and comparing more candidate regions across the context, even where neighboring tokens need not be distinguished. Moreover, attention heads differ in the token relationships they capture, so a selection granularity appropriate for one head may be too coarse or unnecessarily fine for another \citep{correia-etal-2019}.

However, characterizing each head's full input-dependent sensitivity would be costly during inference. Instead, we leverage the distinct, input-independent interaction structure defined by each head's pretrained query--key transformations. This structure identifies the directions to which each head is most responsive, providing a basis for determining both important feature directions and consequential distinctions between context tokens. Along the feature dimension, strongly responsive directions carry the distinctions most consequential to relevance estimation, whereas weak-response directions contribute less and can be represented more coarsely during selection. Along the sequence dimension, the same response structure determines whether tokens need to be distinguished by that head: tokens that differ substantially in the original representation space may induce similar responses if their differences lie primarily along weak-response directions, whereas seemingly similar tokens may need to remain distinct when their differences align with strongly responsive directions. Token redundancy is therefore relative to the attention head rather than solely to the original token representations. Together, these observations motivate jointly adapting feature and sequence resolution according to the same head-specific interaction structure.



Based on this insight, we propose \textbf{HASA}, a training-free sparse attention framework that adapts the resolution used for sparse selection along both feature and sequence dimensions. Along the feature dimension, HASA uses the singular spectrum of each pretrained query--key operator to determine a head-specific effective rank, retaining the dominant interaction directions needed for relevance estimation. Along the sequence dimension, tokens are represented in this compact head-specific space, and neighboring regions are progressively merged when their representations remain directionally coherent, while heterogeneous regions retain finer granularity. These compact representations are used only for routing; exact attention is computed over the original full-dimensional KV states of the selected regions. By reducing both the dimensionality of relevance estimation and the number of candidate regions, HASA lowers selection overhead while retaining head-relevant distinctions for sparse selection. The spectral structure and effective ranks are determined offline, avoiding input-dependent rank-selection overhead during inference.

Our contributions are as follows:
\begin{itemize}
    \item We reveal that each head's pretrained query--key interaction structure characterizes its selection requirements along both feature and sequence dimensions, providing a unified basis for head-adaptive sparse attention.
    \item We develop a head-adaptive feature resolution mechanism that determines each head's effective rank using spectral entropy and a truncation-error criterion, yielding a compact response space that preserves dominant interaction directions.
    \item We develop a head-adaptive sequence resolution mechanism that progressively merges neighboring regions based on token-response coherence in the head-specific response space, yielding candidate regions of varying granularity.
    \item Extensive experiments show that HASA captures \textbf{88.1\%} of dense-attention mass, improves accuracy over UNIQUE by up to \textbf{3.47} and \textbf{2.57} points on RULER and LongBench-Pro, respectively, and achieves self-attention speedups of up to \textbf{14.07$\times$} over dense attention and \textbf{1.32$\times$} over UNIQUE.
\end{itemize}

\section{Related Work}
\label{sec:related_work}
\subsection{Structured Attention and KV-Cache Reduction}
One line of work improves the efficiency of long-context attention through predefined access patterns that restrict which tokens participate in attention. Longformer \citep{beltagy2020longformer} combines local and global attention patterns, while StreamingLLM \citep{xiao2024streamingllm} maintains attention sinks and a bounded recent context for efficient streaming inference. A related line of work reduces the KV cache by retaining selected tokens for future queries. H$_2$O \citep{zhang2023h2o} uses accumulated attention statistics, SnapKV \citep{li2024snapkv} selects positions based on observation-window attention, and DuoAttention \citep{xiao2024duoattention} distinguishes retrieval and streaming heads to reduce cache requirements. These approaches reduce attention computation or memory traffic through structured access or retention decisions. 

However, the regular access patterns and low-overhead selection rules that make these approaches efficient can also constrain their ability to adapt to changing relevance across queries and contexts. HASA retains efficient structured selection through offline head-specific analysis, while dynamically organizing candidate regions and selecting relevant context at inference time.



\subsection{Query-Adaptive Sparse Attention}
Query-adaptive methods determine which tokens or regions to access based on the current query and context. Some approaches learn query-dependent selection explicitly. Native Sparse Attention (NSA) \citep{yuan2025nsa} introduces learned compressed and selected attention branches, SeerAttention \citep{gao2025seerattention} trains a block-level router, and InfLLM-V2 \citep{zhao2025infllmv2} adapts dense checkpoints to sparse retrieval through fine-tuning. SeerAttention-R \citep{gao2026seerattentionr} further extends learned routing to long-horizon reasoning. These approaches introduce additional training or learned selection components. Training-free approaches instead estimate query-dependent relevance at inference time. Quest \citep{tang2024quest} scores fixed KV pages using min--max key statistics, while InfLLM \citep{xiao2024infllm} retrieves relevant context units from external memory. MInference \citep{jiang2024minference}, FlexPrefill \citep{lai2025flexprefill}, and XAttention \citep{xu2025xattention} dynamically infer sparse patterns or block importance, while TokenSelect \citep{wu2025tokenselect}, SpargeAttention \citep{zhang2025spargeattention}, Tactic \citep{zhu2026tactic}, and Evolving Sparsity \citep{han2026evolvingsparsity} reduce selection cost through reuse, filtering, or adaptive sparsification. AB-Sparse \citep{liu2026absparse} and DHSA \citep{DHSA} further consider adaptive block granularity. Among these, UNIQUE \citep{deng2026unique} is closely related to our setting: it represents each fixed KV page using its mean key vector and a scalar dispersion term, and scores relevance using the current full-dimensional query.

However, query-dependent selection alone does not determine the appropriate resolution for each attention head. HASA further accounts for these head-specific differences by using the pretrained query--key interaction structure to adapt both the feature resolution of relevance estimation and the sequence resolution of candidate regions, achieving accurate sparse attention at low cost.

\section{Methodology}
\begin{figure}[t]
    \centering
    \includegraphics[width=\linewidth]{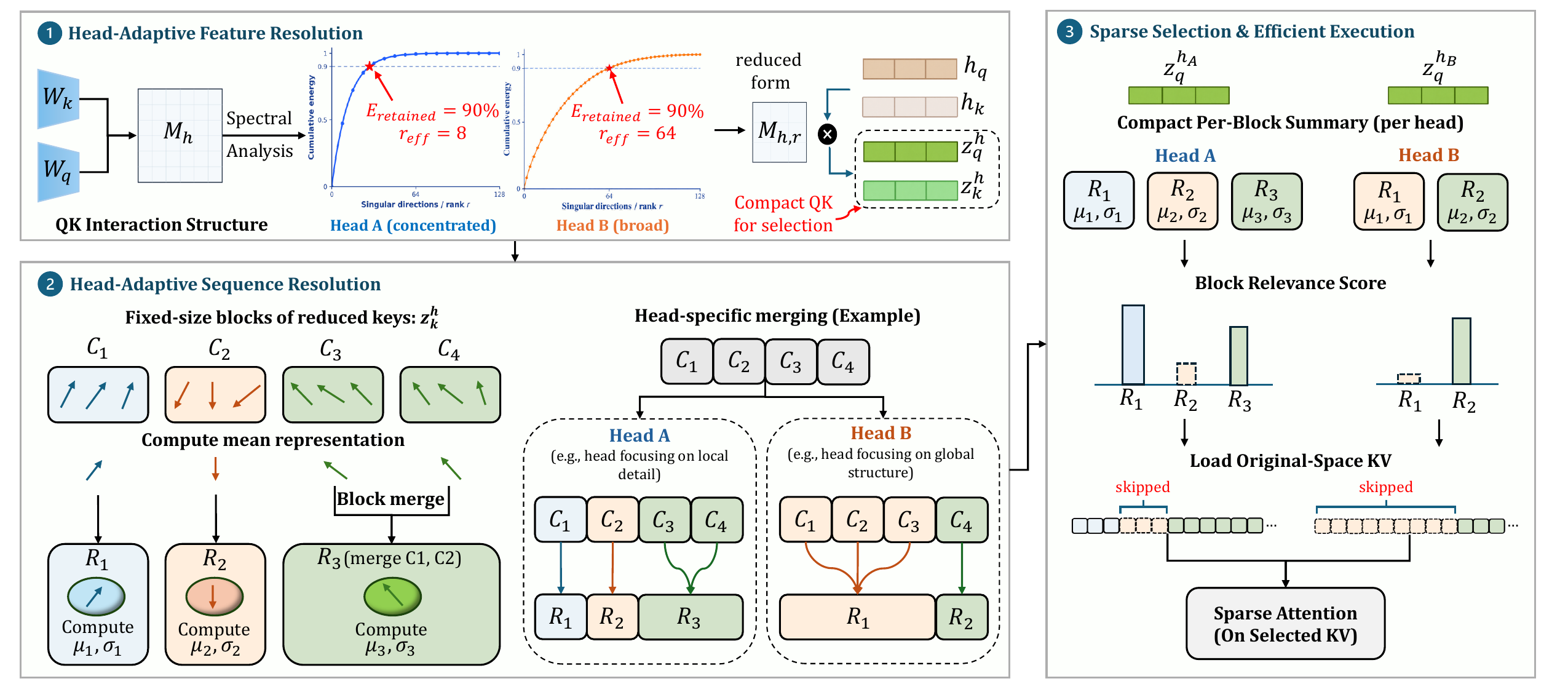}
    \caption{
    \textbf{Overview of HASA.} (1) Each head's pretrained query--key interaction structure determines its effective rank, yielding a compact head-specific response space. (2) In this compact response space, neighboring regions with coherent token responses are progressively merged, while heterogeneous regions remain separate. (3) Relevant regions are selected using compact region summaries, followed by exact attention over their original full-dimensional KV states.
    }
    \label{fig:hasa_overview}
\end{figure}

We present HASA, a training-free sparse attention framework that adapts selection resolution to each head's pretrained query--key interaction structure. As illustrated in Figure~\ref{fig:hasa_overview}, HASA consists of three main stages: (1) head-adaptive feature resolution, which constructs a compact response space for each attention head; (2) head-adaptive sequence resolution, which dynamically organizes candidate context into regions of varying granularity; and (3) sparse selection and efficient execution, which selects relevant regions using compact summaries and computes exact attention over their original keys and values. Together, they enable HASA to achieve accurate and efficient sparse attention.

\paragraph{Problem setup.} Consider an attention head operating on hidden states $\mathbf{H}=[\mathbf{h}_1,\ldots,\mathbf{h}_N]^\top$, where $\mathbf{h}_i\in\mathbb{R}^{d}$. The corresponding query and key representations are $\mathbf{q}_i=\mathbf{W}_Q\mathbf{h}_i$ and $\mathbf{k}_t=\mathbf{W}_K\mathbf{h}_t$, with head dimension $d_h$. Dense attention evaluates $\mathbf{q}_i^\top\mathbf{k}_t/\sqrt{d_h}$ against every preceding key, while sparse attention restricts each query to a selected subset of the candidate context.

\subsection{Head-Adaptive Feature Resolution}
\label{sec:adaptive_feature}

As discussed earlier, sparse selection should account for the query--key interaction structure of each attention head without incurring the substantial overhead of estimating candidate relevance at full feature resolution. We observe that the pretrained query and key projections of each head can be combined into a single bilinear operator. For head $h$, the pre-softmax interaction can be written as

\begin{equation}
    \mathbf{q}_i^\top\mathbf{k}_t
    =
    \mathbf{h}_i^\top
    \underbrace{
        (\mathbf{W}_Q^h)^\top\mathbf{W}_K^h
    }_{\mathbf{M}_h}
    \mathbf{h}_t .
    \label{eq:bilinear_attention}
\end{equation}

The operator $\mathbf{M}_h$ depends only on pretrained model parameters and encodes the head-specific query--key interaction structure. Importantly, the interaction strength need not be distributed uniformly across feature directions: some directions contribute more strongly than others, and this distribution can vary across attention heads. This suggests that different heads may require different numbers of response directions to estimate candidate relevance. To characterize these directions, we consider the singular value decomposition $\mathbf{M}_h=\mathbf{U}_h\boldsymbol{\Sigma}_h\mathbf{V}_h^\top$, with singular values $\sigma_{h,1}\geq\sigma_{h,2}\geq\cdots$. The pre-softmax interaction between hidden states $\mathbf{h}_q$ and $\mathbf{h}_k$ can then be written exactly as

\begin{equation}
    \mathbf{h}_q^\top\mathbf{M}_h\mathbf{h}_k
    =
    \left(
        \boldsymbol{\Sigma}_h^{1/2}
        \mathbf{U}_h^\top\mathbf{h}_q
    \right)^\top
    \left(
        \boldsymbol{\Sigma}_h^{1/2}
        \mathbf{V}_h^\top\mathbf{h}_k
    \right).
    \label{eq:svd_factorization}
\end{equation}

The left and right singular vectors define paired query-side and key-side response directions, while the corresponding singular values characterize their contributions to the bilinear interaction. A concentrated spectrum indicates that the head's interaction structure is dominated by relatively few directions, whereas a flatter spectrum spreads interaction strength across more directions. This motivates determining the retained feature resolution separately for each attention head.

To determine how many response directions to retain for each head, we first normalize the squared singular values as
\[
    p_{h,i}
    =
    \frac{\sigma_{h,i}^2}
         {\sum_j \sigma_{h,j}^2},
\]
where $p_{h,i}$ is the fraction of spectral energy associated with direction $i$. We then compute the spectral entropy and its corresponding effective rank:

\begin{equation}
    H_h
    =
    -\sum_i p_{h,i}\log p_{h,i},
    \qquad
    r_{\mathrm{ent},h}
    =
    \left\lceil \exp(H_h) \right\rceil .
    \label{eq:entropy_rank}
\end{equation}

A concentrated spectrum yields a smaller effective rank, while a flatter spectrum yields a larger one. However, this entropy-based estimate alone does not guarantee that truncation preserves sufficient interaction energy. We therefore complement it with the normalized Frobenius residual of the rank-$r$ approximation $\mathbf{M}_{h,r}$:

\begin{equation}
    \epsilon_{F,h}(r)
    =
    \frac{\|\mathbf{M}_h-\mathbf{M}_{h,r}\|_F^2}
         {\|\mathbf{M}_h\|_F^2}
    =
    \frac{\sum_{i>r}\sigma_{h,i}^2}
         {\sum_i\sigma_{h,i}^2}.
    \label{eq:frobenius_error}
\end{equation}

Given an allowable residual $\epsilon_{\max}$, we define $r_{F,h}=\min\{r:\epsilon_{F,h}(r)\leq\epsilon_{\max}\}$ and select the head-specific rank as
\[
    r_h
    =
    \max\left(r_{\mathrm{ent},h},r_{F,h}\right).
\]
This combines an estimate of the spectrum's effective dimensionality with a constraint on the energy lost through truncation, retaining the principal response directions of each head.

With $r_h$ determined, we retain the leading singular components $\mathbf{U}_{h,r_h}$, $\boldsymbol{\Sigma}_{h,r_h}$, and $\mathbf{V}_{h,r_h}$ to construct paired query-side and key-side response projections:
\begin{equation}
    \mathbf{P}_Q^h
    =
    \boldsymbol{\Sigma}_{h,r_h}^{1/2}
    \mathbf{U}_{h,r_h}^\top,
    \qquad
    \mathbf{P}_K^h
    =
    \boldsymbol{\Sigma}_{h,r_h}^{1/2}
    \mathbf{V}_{h,r_h}^\top .
    \label{eq:response_projection}
\end{equation}
For query and key hidden states, the resulting compact response
representations are
\begin{equation}
    \mathbf{z}_q^h
    =
    \mathbf{P}_Q^h\mathbf{h}_q,
    \qquad
    \mathbf{z}_k^h
    =
    \mathbf{P}_K^h\mathbf{h}_k,
    \label{eq:response_representation}
\end{equation}
which approximate the original query--key interaction through
their inner product:
\begin{equation}
    \mathbf{h}_q^\top\mathbf{M}_h\mathbf{h}_k
    \approx
    (\mathbf{z}_q^h)^\top\mathbf{z}_k^h .
    \label{eq:response_approximation}
\end{equation}

These compact responses enable efficient relevance estimation in a head-adaptive feature space without operating at full feature resolution. The discarded spectrum controls the approximation error of the query--key scores, with the corresponding bound given in Appendix~\ref{app:spectral_bound}. Since $\mathbf{M}_h$ depends only on pretrained model parameters, its spectral decomposition, the head-specific rank $r_h$, and the response projections $\mathbf{P}_Q^h$ and $\mathbf{P}_K^h$ can all be determined offline. HASA thus assigns each head a feature resolution matched to its interaction structure, rather than using the full feature dimension or a fixed rank for all heads.

\subsection{Head-Adaptive Sequence Resolution}
\label{sec:adaptive_sequence}
The head-specific response space constructed above also provides a basis for determining how candidate context should be organized along the sequence dimension. In this space, each token is mapped to a compact key response \(\mathbf z_k^h\) that captures its interaction with the query-side directions retained by head \(h\). Neighboring tokens with coherent key responses can therefore be summarized together, whereas tokens with heterogeneous responses should remain distinguishable for subsequent sparse selection. HASA uses local response coherence to construct candidate regions at varying granularities, adapting sequence resolution to each head's interaction structure.


To organize head-specific key responses into candidate regions, we first define how a group of neighboring tokens is summarized for sparse selection. To reduce candidate evaluation cost, HASA represents a candidate region $C$ through a shared summary of its compact key responses rather than scoring each token individually. For clarity, we omit the attention-head superscript below and write $\mathbf{z}_i\in\mathbb{R}^{r_h}$ for the compact key response of token $i$. We define the region's mean response as

\begin{equation}
    \boldsymbol{\mu}_C
    =
    \frac{1}{|C|}
    \sum_{i\in C}\mathbf{z}_i ,
    \label{eq:block_mean}
\end{equation}

and its scalar dispersion as

\begin{equation}
    \sigma_C
    =
    \sqrt{
        \frac{1}{|C|}
        \sum_{i\in C}
        \|\mathbf{z}_i-\boldsymbol{\mu}_C\|_2^2
    }.
    \label{eq:block_dispersion}
\end{equation}

Following the mean--dispersion scoring formulation of \cite{deng2026unique}, we define the relevance of region $C$ to a compact query response $\mathbf{z}_q$ as

\begin{equation}
    s(\mathbf{z}_q,C)
    =
    \mathbf{z}_q^\top\boldsymbol{\mu}_C
    +
    \lambda\|\mathbf{z}_q\|_2\sigma_C ,
    \label{eq:semantic_block_score}
\end{equation}

where $\lambda$ controls the contribution of within-region dispersion and is fixed to $0.5$ in all experiments. The mean term captures the region's average interaction with the query, while the dispersion term accounts for variation that may be suppressed by mean pooling.

After defining the region-level selection representation, we progressively merge neighboring regions whose key-response directions remain coherent. Here, comparing only region means is insufficient, as similarly oriented means may conceal heterogeneous token responses within a region. We therefore assess directional coherence across the combined region using the normalized key responses $\hat{\mathbf{z}}_i =\mathbf{z}_i/(\|\mathbf{z}_i\|_2)$. For a region $C$, we define

\begin{equation}
    \mathbf{R}_C
    =
    \sum_{i\in C}\hat{\mathbf{z}}_i,
    \qquad
    \rho_C
    =
    \frac{\|\mathbf{R}_C\|_2}{|C|},
    \label{eq:block_purity}
\end{equation}

A high $\rho_C$ indicates concentrated response directions, whereas a low value indicates greater directional heterogeneity. This measure is related to the average pairwise agreement among token response directions and can be computed from the additive resultant $\mathbf{R}_C$; the derivation is provided in Appendix~\ref{app:purity_derivation}.

We initialize the sequence with contiguous base regions of $S_0$ tokens ($S_0=8$ in our implementation) and construct larger regions through incremental merging. This bottom-up procedure allows both merge decisions and selection summaries to be updated from additive region statistics, without revisiting the constituent tokens after initialization. For each active region $C$, we maintain
\[
    n_C=|C|,
    \qquad
    \mathbf{R}_C=\sum_{i\in C}\hat{\mathbf{z}}_i,
    \qquad
    \mathbf{S}_C=\sum_{i\in C}\mathbf{z}_i,
    \qquad
    Q_C=\sum_{i\in C}\|\mathbf{z}_i\|_2^2.
\]
For two adjacent regions $A$ and $B$, the directional coherence of their union is
\begin{equation}
    \rho_{A\cup B}
    =
    \frac{\|\mathbf{R}_A+\mathbf{R}_B\|_2}
         {n_A+n_B}.
    \label{eq:merge_purity}
\end{equation}
We merge $A$ and $B$ when $\rho_{A\cup B}\geq\rho_{\min}$, subject to the maximum region-size constraint $n_A+n_B\leq C_{\max}$. The additive statistics also allow the selection summary of a merged region $C=A\cup B$ to be updated without revisiting its constituent tokens:

\begin{equation}
    \boldsymbol{\mu}_C
    =
    \frac{\mathbf{S}_C}{n_C},
    \qquad
    \sigma_C
    =
    \sqrt{
        \frac{Q_C}{n_C}
        -
        \|\boldsymbol{\mu}_C\|_2^2
    }.
    \label{eq:merged_block_statistics}
\end{equation}

Since the merge criterion and summary updates use only additive region statistics, each candidate merge requires $O(r_h)$ operations, independent of the number of tokens in the regions. This incremental procedure produces head-adaptive candidate regions at varying granularities, each equipped with a compact summary for subsequent sparse selection via~\eqref{eq:semantic_block_score}.

\subsection{Sparse Selection and Efficient Execution}
\label{sec:efficient_execution}

After constructing the head-adaptive candidate regions, HASA uses their compact summaries to determine which parts of the context should be accessed for each query. For head $h$, the current query is projected into its head-specific response space to obtain $\mathbf{z}_q^h$. We then score each candidate region $C$ using its maintained summary $(\boldsymbol{\mu}_C,\sigma_C)$ according to~\eqref{eq:semantic_block_score}. Let $I$ denote the number of base regions and $I_h$ the number of candidate regions after head-adaptive merging for head $h$. Scoring these regions requires $O(I_h r_h)$ operations per query, compared with $O(Id)$ for scoring the base regions at full feature resolution. Since merging does not increase the number of regions and $r_h \leq d$, head-adaptive sequence and feature resolutions jointly reduce the region-scoring cost. This produces a query-dependent ranking of candidate regions without computing full-dimensional attention scores for their constituent tokens.

HASA selects the top-$K$ regions according to these scores and additionally includes the local context. Let $\mathcal{C}_q^h$ denote the selected candidate regions and $\mathcal{L}_q$ the local token positions. Together, they define the sparse attention support:
\begin{equation}
    \mathcal{I}_q^h
    =
    \left(
        \cup_{C\in\mathcal{C}_q^h} C
    \right)
    \cup \mathcal{L}_q.
    \label{eq:sparse_support}
\end{equation}

Once $\mathcal{I}_q^h$ is determined, HASA retrieves the original full-dimensional keys and values at the selected positions and computes exact attention over this sparse region. By scoring compact region summaries rather than individual full-dimensional keys, HASA keeps the selection lightweight. By accessing only the selected KV, it also reduces memory traffic and attention computation.

Overall, HASA combines head-adaptive feature and sequence resolutions to identify relevant context at an appropriate granularity, while using the original query, keys, and values for attention over the selected tokens. This design preserves attention accuracy while reducing inference latency.


\begin{table}[t]
\centering
\caption{Accuracy results on LongBench-Pro using Ministral-3-8B-Instruct-2512. The budget denotes the number of KV tokens available to each query during decoding; full attention is unrestricted.}
\label{tab:longbench}
\fontsize{8}{7.5}\selectfont
\setlength{\tabcolsep}{3pt}{
\begin{tabular}{l|c|c|cc|cccc}
\toprule
\textbf{Method} & \textbf{Budget} &
\textbf{Overall} & \textbf{En} & \textbf{Zh} &
\textbf{Extreme} & \textbf{Hard} &
\textbf{Moderate} & \textbf{Easy} \\
\midrule

Full attention & -- &
37.70 & 35.88 & 38.96 & 31.13 & 33.72 & 31.60 & 48.86 \\

\midrule

Quest & 2048 &
30.63 & 29.17 & 32.60 & 24.60 & 28.60 & 24.17 & 42.01 \\

H2O & 2048 &
33.02 & 31.77 & 34.42 & 27.51 & 31.08 & 26.56 & 43.33 \\

InfLLM & 2048 &
36.16 & 35.06 & 37.53 & 29.95 & 32.90 & 29.84 & \textbf{47.98} \\

UNIQUE & 2048 &
37.11 & 35.57 & 38.15 & 31.28 & 33.18 & \textbf{31.03} & 47.66 \\

\textbf{HASA (Ours)} & 2048 &
\textbf{37.62} & \textbf{36.90} & \textbf{38.28} & \textbf{32.80} & \textbf{34.10} & 31.01 & 47.25 \\

\midrule

Quest & 1024 &
27.18 & 25.83 & 28.76 & 21.53 & 22.69 & 21.62 & 38.71 \\

H2O & 1024 &
31.61 & 30.59 & 33.16 & 26.93 & 29.15 & 25.43 & 41.87 \\

InfLLM & 1024 &
35.42 & 34.77 & 36.68 & 29.21 & 32.58 & 28.60 & 48.12 \\

UNIQUE & 1024 &
36.72 & 35.77 & 37.98 & 31.02 & 32.78 & 30.75 & 48.34 \\

\textbf{HASA (Ours)} & 1024 &
\textbf{37.53} & \textbf{36.66} & \textbf{38.74} &
\textbf{32.17} & \textbf{33.83} & \textbf{32.27} & \textbf{48.63} \\

\midrule

Quest & 512 &
21.72 & 21.09 & 21.84 & 18.87 & 16.18 & 17.61 & 29.31 \\

H2O & 512 &
29.04 & 27.96 & 30.44 & 23.61 & 28.15 & 22.20 & 39.13 \\

InfLLM & 512 &
34.99 & 33.82 & 36.21 & 28.96 & 32.38 & 27.31 & 46.74 \\

UNIQUE & 512 &
36.58 & \textbf{35.22} & 38.23 & 30.86 & 33.75 & 30.37 & 47.67 \\

\textbf{HASA (Ours)} & 512 &
\textbf{37.43} & 35.02 & \textbf{40.13} &
\textbf{30.91} & \textbf{34.27} & \textbf{32.94} & \textbf{48.46} \\

\bottomrule
\end{tabular}
}
\end{table}

\section{Experiments}

\subsection{Experimental Setup}
We evaluate HASA on \textbf{LongBench-Pro} \citep{longbench} and \textbf{RULER-32K} \citep{ruler}, covering realistic multilingual long-context tasks and controlled synthetic tasks, respectively. Experiments are conducted on \textbf{Ministral-3-8B-Instruct-2512} and \textbf{Qwen3-8B} to assess generalization across model families. We compare HASA with four training-free sparse-attention baselines: \textbf{H$_2$O} \citep{zhang2023h2o}, \textbf{Quest} \citep{tang2024quest}, \textbf{InfLLM} \citep{xiao2024infllm}, and \textbf{UNIQUE} \citep{deng2026unique}, together with \textbf{dense attention} as the full-context reference. Sparse methods are evaluated under matched per-query KV budgets across multiple sparsity levels. We report overall, language-specific, and difficulty-level scores on LongBench-Pro and the average score on RULER-32K. Benchmark details, baseline configurations, and evaluation protocols are in Appendix~\ref{app:setup}.

\begin{table*}[t]
\centering
\caption{Accuracy results on RULER-32K using Ministral-3-8B-Instruct-2512. The budget denotes the number of KV tokens available to each query during decoding; full attention is unrestricted.}
\label{tab:ruler}
\resizebox{\textwidth}{!}{
\begin{tabular}{l|c|ccccccccccccc|c}
\toprule
\textbf{Method} & \textbf{Budget} &
\textbf{SG1} & \textbf{SG2} & \textbf{SG3} & \textbf{MK1} &
\textbf{MK2} & \textbf{MK3} & \textbf{MV} &
\textbf{MQ} & \textbf{VT} & \textbf{CWE} &
\textbf{FWE} & \textbf{QA1} & \textbf{QA2} &
\textbf{Avg.} \\
\midrule

Full attention & -- &
100.00 & 99.60 & 99.60 & 99.40 & 100.00 & 99.80 & 98.55 &
99.05 & 99.96 & 88.46 & 93.27 & 75.80 & 57.40 & 93.15 \\

\midrule

Quest & 2048 &
\textbf{100.00} & 98.80 & 97.00 & 99.20 & 99.80 & 97.60 & 99.45 &
99.30 & 99.88 & 77.80 & 90.67 & 75.00 & 57.20 & 91.67 \\

H2O & 2048 &
\textbf{100.00} & 99.80 & 31.00 & 99.60 & 97.80 & 1.80 & 92.65 &
65.20 & 98.40 & 42.96 & 90.80 & 74.40 & 56.80 & 73.17 \\

InfLLM & 2048 &
\textbf{100.00} & 94.00 & 85.20 & 95.20 & 99.80 & 96.40 & 93.00 &
88.45 & 94.08 & 22.10 & 83.80 & 74.60 & 57.20 & 83.37 \\

UNIQUE & 2048 &
\textbf{100.00} & 99.80 & 99.20 & 99.60 & \textbf{100.00} & 99.80 & \textbf{99.00} &
\textbf{99.15} & 99.96 & 82.28 & \textbf{91.07} & 75.60 & \textbf{57.40} & 92.53 \\

\textbf{HASA (Ours)} & 2048 &
\textbf{100.00} & \textbf{100.00} & \textbf{100.00} & \textbf{100.00} & 99.00 & \textbf{99.80} & 98.51 &
\textbf{99.15} & \textbf{100.00} & \textbf{83.86} & 90.43 & \textbf{76.53} & \textbf{57.40} & \textbf{92.67} \\

\midrule

Quest & 1024 &
\textbf{100.00} & 98.80 & 97.00 & 99.16 & 99.76 & 86.45 & \textbf{99.44} &
\textbf{99.29} & 99.88 & 71.61 & 90.53 & 74.86 & 56.72 & 90.27 \\

H2O & 1024 &
99.90 & 98.85 & 17.94 & 98.72 & 90.67 & 0.92 & 67.59 &
41.44 & 82.32 & 32.39 & 79.12 & 74.40 & 56.31 & 64.66 \\

InfLLM & 1024 &
96.52 & 89.68 & 80.78 & 92.23 & 99.16 & 69.57 & 89.52 &
84.02 & 90.67 & 16.35 & 79.01 & 74.41 & 56.74 & 78.36 \\

UNIQUE & 1024 &
\textbf{100.00} & 99.80 & 99.20 & 99.60 & \textbf{100.00} & \textbf{99.80} &
99.00 & 99.15 & 99.96 & 75.38 & \textbf{90.96} & 75.16 & \textbf{57.56} & 91.97 \\

HASA (Ours) & 1024 &
\textbf{100.00} & \textbf{100.00} & \textbf{100.00} & \textbf{100.00} &
99.01 & 99.79 & 98.51 & 99.15 & \textbf{100.00} & \textbf{78.39} &
90.27 & \textbf{76.09} & \textbf{57.56} & \textbf{92.21} \\

\midrule

Quest & 512 &
99.40 & 94.60 & 91.00 & 97.20 & 95.80 & 53.00 & 96.55 &
96.65 & 99.00 & 53.24 & 87.60 & 74.00 & 55.80 & 84.14 \\

H2O & 512 &
99.80 & 98.00 & 2.60 & 97.60 & 84.60 & 0.00 & 44.45 &
14.65 & 66.88 & 20.94 & 68.33 & 74.40 & 55.80 & 56.00 \\

InfLLM & 512 &
92.60 & 86.00 & 76.00 & 88.60 & 98.40 & 43.80 & 85.60 &
79.40 & 87.52 & 10.12 & 74.40 & 74.20 & 56.20 & 73.30 \\

UNIQUE & 512 &
\textbf{100.00} & 99.80 & 99.20 & \textbf{99.80} & \textbf{100.00} & \textbf{99.60} & 99.55 &
\textbf{99.30} & 99.96 & 62.90 & \textbf{88.67} & 73.60 & \textbf{57.80} & 90.78 \\

\textbf{HASA (Ours)} & 512 &
\textbf{100.00} & \textbf{100.00} & \textbf{100.00} & \textbf{99.80} & \textbf{100.00} & 99.20 & \textbf{100.00} & \textbf{100.00} & \textbf{100.00} & \textbf{66.37} & 87.73 & \textbf{74.54} & 57.40 & \textbf{91.16} \\

\bottomrule
\end{tabular}
}
\end{table*}

\begin{figure*}[t]
    \centering
    \includegraphics[width=\textwidth]{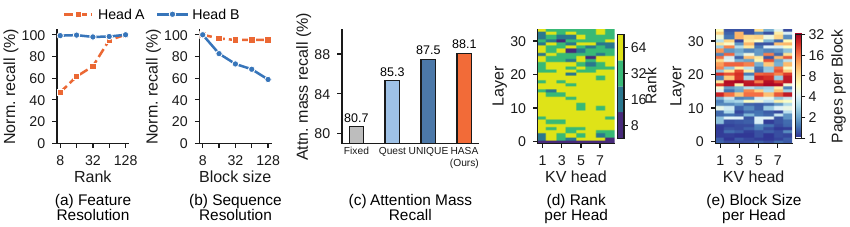}
    \caption{
    \textbf{Head-specific resolution and attention recall.} (a--b) High recall is achievable by adjusting feature rank and block size, with different heads requiring different resolutions. (c) Adaptive resolution improves normalized attention recall over fixed configurations across RULER-32K tasks. (d--e) HASA selects varying feature ranks and sequence granularities across layers and KV heads, confirming head-specific adaptation.
    }
    \label{fig:adaptive_resolution_analysis}
\end{figure*}


\subsection{Results and Analysis}
\paragraph{Accuracy.}
Tables~\ref{tab:longbench} and~\ref{tab:ruler} compare HASA with the baselines under matched KV budgets on Ministral-3-8B-Instruct-2512. Across both benchmarks, HASA achieves the highest overall scores among the evaluated sparse methods, with scores close to those of dense attention.

\begin{wrapfigure}{r}{0.49\columnwidth}
    \centering
    \begin{subfigure}[t]{0.48\linewidth}
        \centering
        \includegraphics[width=\linewidth]{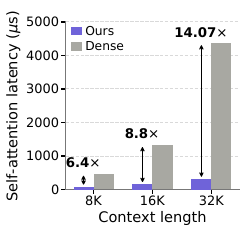}
        \caption{Comparison with dense attention.}
        \label{fig:latency_dense}
    \end{subfigure}
    \hfill
    \begin{subfigure}[t]{0.48\linewidth}
        \centering
        \includegraphics[width=\linewidth]{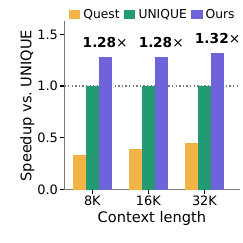}
        \caption{Comparison with sparse baselines.}
        \label{fig:latency_sparse}
    \end{subfigure}

    \caption{
    Self-attention efficiency at increasing context lengths. 
    \textbf{(a)} Our method substantially reduces latency compared with dense attention.
    \textbf{(b)} Our method achieves higher self-attention speedup than Quest and UNIQUE.
    }
    \label{fig:latency}
\end{wrapfigure}

On LongBench-Pro, HASA achieves \textbf{37.62}, \textbf{37.53}, and \textbf{37.43} at KV budgets of 2048, 1024, and 512, respectively, compared with 37.11, 36.72, and 36.58 for UNIQUE. The advantage becomes more pronounced as the budget decreases: at 512 tokens, HASA remains within only 0.27 points of dense attention (37.70). The gains also extend across language and difficulty breakdowns. At a 1024-token budget, HASA scores \textbf{36.66} on English and \textbf{38.74} on Chinese, with strong performance on the Extreme, Hard, and Moderate subsets.

On RULER-32K, HASA obtains average scores of \textbf{92.67}, \textbf{92.21}, and \textbf{91.16} at the same budgets, compared with 92.53, 91.97, and 90.78 for UNIQUE. HASA also maintains strong performance on retrieval and multi-key tasks, including challenging settings such as CWE and QA1. Reducing the budget from 2048 to 512 lowers HASA's average score by only 1.51 points. Together, these results suggest that head-adaptive selection helps retain relevant context when the available KV budget becomes increasingly restrictive. Similar trends are observed on \textbf{Qwen3-8B} (Appendix~\ref{app:additional_eval}).

\paragraph{Adaptive Resolution.}
Figure~\ref{fig:adaptive_resolution_analysis} examines the effectiveness of head-adaptive feature and sequence resolutions. Figures~\ref{fig:adaptive_resolution_analysis}(a--b) reveal distinct resolution requirements across heads: Head B retains \textbf{99.2\%} recall at rank 8, while Head A requires rank 64 to reach \textbf{95.1\%}; conversely, at block size 128, Head A maintains \textbf{95.2\%} recall whereas Head B drops to \textbf{59\%}. These contrasting requirements motivate joint adaptation across both dimensions. Figure~\ref{fig:adaptive_resolution_analysis}(c) evaluates the
resulting selection fidelity using attention-mass recall under
matched KV budgets, where the full dense-attention mass is 100\%. For the fixed-resolution reference, we disable both forms of head-wise adaptation: all heads use a shared feature rank equal to the average of HASA's head-specific ranks and fixed 8-token pages without adaptive merging. HASA achieves \textbf{88.1\%} attention-mass recall, compared with \textbf{80.7\%} for this fixed configuration, an improvement of \textbf{7.4 percentage points. It also exceeds Quest (85.3\%) and UNIQUE (87.5\%), indicating that adapting selection resolution better preserves the context emphasized by dense attention.} Finally, Figures~\ref{fig:adaptive_resolution_analysis}(d--e) visualize the varying feature and sequence resolutions across layers and KV heads, confirming HASA's adaptation to the underlying head interaction structure.

\paragraph{Efficiency.}
Figure~\ref{fig:latency} reports self-attention latency across increasing context lengths. HASA achieves speedups of \textbf{6.4$\times$}, \textbf{8.8$\times$}, and \textbf{14.07$\times$} over dense attention at 8K, 16K, and 32K, respectively, demonstrating increasing benefits from sparse execution at longer contexts. HASA also improves over the training-free sparse-attention baselines: compared with UNIQUE, it achieves \textbf{1.28$\times$}, \textbf{1.28$\times$}, and \textbf{1.32$\times$} speedups at the same context lengths. This additional gain over a sparse baseline is consistent with HASA's use of compact head-specific responses and region summaries to reduce query-dependent selection cost, with the advantage becoming more pronounced for longer context.

\begin{wraptable}{r}{0.43\columnwidth}
\vspace{-13pt}
\centering
\caption{Ablation of adaptive feature and sequence resolution at 1024 KV budget, RULER-32K.}
\label{tab:ablation_rank_block}

\small
\setlength{\tabcolsep}{3pt}
\resizebox{\linewidth}{!}{%
\begin{tabular}{cc|c|c}
\toprule
\textbf{Feature} &
\textbf{Sequence} &
\textbf{Acc. (\%)} &
\makecell[c]{\textbf{Norm.}\\\textbf{Recall (\%)}} \\
\midrule
Fixed    & Fixed    & 91.18 & 91.60 \\
Adaptive & Fixed    & 91.42 & 93.40 \\
Fixed    & Adaptive & 91.50 & 94.00 \\
\textbf{Adaptive} & \textbf{Adaptive} & \textbf{92.21} & \textbf{100.00} \\
\bottomrule
\end{tabular}
}
\end{wraptable}

\paragraph{Ablation Study.}


Table~\ref{tab:ablation_rank_block} isolates the contributions of adaptive feature and sequence resolution at a 1024-token KV budget on RULER-32K. The fully fixed configuration follows the fixed-resolution reference in Figure~\ref{fig:adaptive_resolution_analysis}(c), using a shared feature rank and fixed 8-token pages without adaptive merging. Starting from the fully fixed configuration, adaptive feature resolution improves accuracy from 91.18\% to 91.42\% and normalized attention recall from 91.60\% to 93.40\%, while adaptive sequence resolution achieves 91.50\% accuracy and 94.00\% recall. Combining both yields the highest accuracy and normalized recall, reaching 92.21\% and 100\%, respectively, corresponding to gains of 1.03 and 8.40 percentage points over the fully fixed configuration. These results show that the two adaptations are complementary: each individually improves preservation of dense-attention relevance, while their combination further improves both selection fidelity and downstream accuracy.

\section{Conclusion}
\label{sec:conclusion}
We presented \textbf{HASA}, a training-free sparse attention framework that adapts feature and sequence resolutions to the pretrained query--key interaction structure of each attention head. HASA determines head-specific feature ranks through spectral analysis, retaining the dominant interaction directions. It further constructs candidate regions according to the coherence of compact key responses, adapting sequence granularity to each head's interaction structure. HASA then performs exact attention over the selected full-dimensional keys and values. Across benchmarks, KV budgets, and model backbones, HASA improves long-context accuracy over training-free sparse-attention baselines while remaining close to dense attention. Quantitatively, HASA captures \textbf{88.1\%} of dense-attention mass, improves accuracy over UNIQUE by up to \textbf{3.47 points} on RULER and \textbf{2.57 points} on LongBench-Pro, and accelerates self-attention by up to \textbf{14.07$\times$} over dense attention and \textbf{1.32$\times$} over UNIQUE.

%


\paragraph{Limitations.} HASA characterizes each attention head using the pretrained bilinear query--key operator $\mathbf{M}_h=(\mathbf{W}_Q^h)^\top\mathbf{W}_K^h$, enabling its response structure to be determined offline. However, this input-independent characterization does not capture the full runtime sensitivity of attention, including the effects of softmax normalization, value representations, and downstream computation. Input-dependent information, such as local Jacobians, could provide a more faithful characterization but would incur additional inference overhead. Future work could explore lightweight input-dependent corrections that improve selection fidelity while preserving HASA's efficiency.

\subsection*{AI Usage Statement}
In this work, Generative AI was used to improve the grammar, clarity, and readability of the author-written text, and \LaTeX{} formatting issues. Coding assistants were also used for implementation tasks, such as generating plots, improving code organization and structure, and assisting with the routine implementation. Generative AI was not used to develop research ideas, formulate hypotheses, design experiments, analyze experimental results, or draw scientific conclusions. All AI-assisted text and code were carefully reviewed by the authors, and generated code was verified and tested for correctness before use. The authors take full responsibility for the final content of this work, including all text, claims, experimental results, code, and artifacts produced with the assistance of generative AI.

\subsection*{Ethics Statement}
This work uses publicly available models and benchmarks and involves no human subjects, personally identifiable information, or new data collection.

\bibliography{main}
\bibliographystyle{iclr2027_conference}

\clearpage
\appendix
\section{Theoretical Grounding of Spectral Routing}
\label{app:spectral_bound}

\paragraph{Approximation of region scores.}
HASA ranks candidate regions using the mean--dispersion score in~\eqref{eq:semantic_block_score}, rather than scoring individual keys. We therefore examine how spectral truncation affects this region-level score.

For a fixed region $C$, let $S_C^{(p)}$ and $S_C^{(r)}$ denote the scores computed from the full and rank-$r$ spectral response representations, respectively, using the same mean--dispersion formulation. Here $p=\operatorname{rank}(M_h)$, so the full response retains all nonzero singular directions. Define the mean and dispersion of the original key hidden states within $C$ as
\[
\bar{x}_C
=
\frac{1}{|C|}\sum_{i\in C}x_{k,i},
\qquad
v_C
=
\left(
\frac{1}{|C|}
\sum_{i\in C}
\|x_{k,i}-\bar{x}_C\|_2^2
\right)^{1/2}.
\]

\begin{lemma}[Region-score approximation]
\label{lem:block_score_approx}
Let $\gamma=\sigma_{h,r+1}$ and $L=\sigma_{h,1}$. For any fixed query and region $C$, the mean--dispersion scores satisfy
\[
\left|S_C^{(p)}-S_C^{(r)}\right|
\leq
\gamma\|x_q\|_2\|\bar{x}_C\|_2
+
\lambda\sqrt{2L\gamma}\,
\|x_q\|_2v_C.
\]
\end{lemma}

\begin{proof}
Partition each full spectral response into its retained and discarded components. Because these components occupy orthogonal coordinates, the difference between the full and truncated mean terms is
\[
\begin{aligned}
\left|
(\mathbf z_q^{(p)})^\top\boldsymbol{\mu}_C^{(p)}
-
(\mathbf z_q^{(r)})^\top\boldsymbol{\mu}_C^{(r)}
\right|
&=
\left|
x_q^\top(M_h-M_{h,r})\bar{x}_C
\right|\\
&\leq
\gamma\|x_q\|_2\|\bar{x}_C\|_2.
\end{aligned}
\]

It remains to bound the dispersion contribution. Let $a_p=\|\mathbf z_q^{(p)}\|_2$ and $a_r=\|\mathbf z_q^{(r)}\|_2$, and let $\sigma_C^{(p)}$ and $\sigma_C^{(r)}$ denote the corresponding response dispersions. Orthogonality of the retained and discarded components gives
\[
a_p^2-a_r^2
\leq \gamma\|x_q\|_2^2,
\qquad
(\sigma_C^{(p)})^2-(\sigma_C^{(r)})^2
\leq \gamma v_C^2.
\]
Moreover,
\[
a_r^2\leq L\|x_q\|_2^2,
\qquad
(\sigma_C^{(p)})^2\leq Lv_C^2.
\]
Since $a_p\geq a_r$ and
$\sigma_C^{(p)}\geq\sigma_C^{(r)}$,
\[
\begin{aligned}
&\left|
a_p\sigma_C^{(p)}
-
a_r\sigma_C^{(r)}
\right|^2\\
&\quad\leq
a_p^2(\sigma_C^{(p)})^2
-
a_r^2(\sigma_C^{(r)})^2\\
&\quad=
(a_p^2-a_r^2)(\sigma_C^{(p)})^2
+
a_r^2\left(
(\sigma_C^{(p)})^2
-
(\sigma_C^{(r)})^2
\right)\\
&\quad\leq
2L\gamma\|x_q\|_2^2v_C^2.
\end{aligned}
\]
Multiplying by $\lambda$ and combining the bounds for the mean and dispersion terms proves the result.
\end{proof}

\paragraph{Preservation of block selection.}
A bounded region-score error does not guarantee identical ranking when candidate regions have nearly identical scores. However, a sufficient margin at the selection boundary preserves the selected blocks.

\begin{corollary}[Top-$K$ block selection preservation]
\label{cor:topk}
Consider a fixed query and a fixed collection of candidate regions $\mathcal{C}=\{C_1,\ldots,C_n\}$. Let $S_j^{(p)}$ and $S_j^{(r)}$ be their full-rank and rank-$r$ mean--dispersion scores, respectively, and let
\[
\epsilon
=
\max_{1\leq j\leq n}
\left[
\gamma\|x_q\|_2\|\bar{x}_{C_j}\|_2
+
\lambda\sqrt{2L\gamma}\,
\|x_q\|_2v_{C_j}
\right].
\]
If the $K$-th and $(K+1)$-th largest full-rank region scores satisfy
\[
S_{(K)}^{(p)}-S_{(K+1)}^{(p)}>2\epsilon,
\]
then the top-$K$ region sets under the full-rank and rank-$r$ scores are identical.
\end{corollary}

\begin{proof}
Lemma~\ref{lem:block_score_approx} gives $|S_j^{(p)}-S_j^{(r)}|\leq\epsilon$ for every candidate region. For any region $C_i$ in the full-rank top-$K$ set and any region $C_j$ outside it,
\[
S_i^{(r)}
\geq S_{(K)}^{(p)}-\epsilon,
\qquad
S_j^{(r)}
\leq S_{(K+1)}^{(p)}+\epsilon.
\]
The assumed margin therefore implies $S_i^{(r)}>S_j^{(r)}$ for every such pair. Hence no region outside the full-rank top-$K$ set can displace a selected region.
\end{proof}

Together, these results clarify the theoretical role of spectral truncation in HASA. The discarded singular spectrum bounds both the underlying query--key interaction error and the approximation error of the mean--dispersion region scores used for routing. Under a sufficient selection margin, this bound further guarantees preservation of top-$K$ selection over a fixed collection of candidate regions. The analysis isolates the effect of feature truncation and does not guarantee identical region partitions when feature rank changes. After selection, HASA computes exact attention over the selected original full-dimensional KV states.

\section{Analysis of the Directional Purity Criterion}
\label{app:purity_derivation}

This section provides additional analysis of the directional-purity criterion introduced in Section~\ref{sec:adaptive_sequence}. In particular, we establish its relationship to pairwise directional agreement and contrast it with directly comparing neighboring block representatives.

\subsection{Relation to Pairwise Directional Agreement}

Recall that for a block $C$ with $n_C=|C|$ normalized nonzero token responses $\hat{\mathbf{z}}_i$, directional purity is defined as
\[
    \rho_C =
    \frac{1}{n_C}
    \left\|
        \sum_{i\in C}\hat{\mathbf{z}}_i
    \right\|_2 .
\]
Squaring both sides gives
\begin{align}
    \rho_C^2
    &=
    \frac{1}{n_C^2}
    \left\|
        \sum_{i\in C}\hat{\mathbf{z}}_i
    \right\|_2^2 \\
    &=
    \frac{1}{n_C^2}
    \left(
        \sum_{i\in C}\|\hat{\mathbf{z}}_i\|_2^2
        +
        2\sum_{i<j}
        \hat{\mathbf{z}}_i^\top\hat{\mathbf{z}}_j
    \right).
\end{align}
Since $\|\hat{\mathbf{z}}_i\|_2=1$, this reduces to
\begin{equation}
    \rho_C^2
    =
    \frac{1}{n_C^2}
    \left(
        n_C +
        2\sum_{i<j}
        \hat{\mathbf{z}}_i^\top\hat{\mathbf{z}}_j
    \right).
    \label{eq:purity_pairwise_app}
\end{equation}

Because the representations are normalized,
$\hat{\mathbf{z}}_i^\top\hat{\mathbf{z}}_j$ is their cosine similarity.
For $n_C\geq 2$, let the average pairwise cosine agreement within $C$ be
\[
    \bar{c}_C =
    \frac{2}{n_C(n_C-1)}
    \sum_{i<j}
    \hat{\mathbf{z}}_i^\top\hat{\mathbf{z}}_j .
\]
Substituting into ~\eqref{eq:purity_pairwise_app} yields
\begin{equation}
    \rho_C^2
    =
    \frac{1}{n_C}
    +
    \frac{n_C-1}{n_C}\bar{c}_C,
\end{equation}
and equivalently,
\begin{equation}
    \bar{c}_C
    =
    \frac{n_C\rho_C^2-1}{n_C-1}.
    \label{eq:purity_cosine_relation_app}
\end{equation}

Thus, $\rho_C$ is directly related to the average pairwise directional
agreement of all token representations within the block. Importantly, it can
be computed from a single resultant vector, avoiding explicit evaluation of
the $\mathcal{O}(n_C^2)$ token pairs.

\subsection{Why Block-Mean Cosine Is Insufficient}
An alternative merge criterion is to compare neighboring block representatives using $\cos(\boldsymbol{\mu}_A,\boldsymbol{\mu}_B)$. This quantity, however, depends only on the two means and contains no information about the distribution of directions within either block.

To see this, consider two blocks whose token representations are internally dispersed but whose mean directions are similar. Their block means may satisfy $\cos(\boldsymbol{\mu}_A,\boldsymbol{\mu}_B) \approx 1$, even though the tokens within $A\cup B$ exhibit weak directional agreement. A cosine-based merge would therefore treat the blocks as compatible solely because their first-order summaries align.

Directional purity instead evaluates
\[
    \rho_{A\cup B}
    =
    \frac{\|\mathbf{R}_A+\mathbf{R}_B\|_2}
         {n_A+n_B},
\]
which depends on the accumulated directions of all tokens in the proposed
union. As shown by ~\eqref{eq:purity_cosine_relation_app}, a high
$\rho_{A\cup B}$ implies high aggregate pairwise agreement within the merged
region, rather than merely alignment between two block means.

\subsection{Exact Update of Pooled Statistics}
The additive statistics maintained during merging also allow the pooled
quantities used for block scoring to be recovered without revisiting
individual tokens. For a block $C$, let
\[
    \mathbf{S}_C = \sum_{i\in C}\mathbf{z}_i,
    \qquad
    Q_C = \sum_{i\in C}\|\mathbf{z}_i\|_2^2 .
\]
The mean follows immediately as
\[
    \boldsymbol{\mu}_C = \frac{\mathbf{S}_C}{n_C}.
\]
Using the identity
\[
    \frac{1}{n_C}
    \sum_{i\in C}
    \|\mathbf{z}_i-\boldsymbol{\mu}_C\|_2^2
    =
    \frac{Q_C}{n_C}
    -
    \|\boldsymbol{\mu}_C\|_2^2,
\]
the scalar dispersion used by the block-scoring function is
\[
    \sigma_C
    =
    \sqrt{
        \frac{Q_C}{n_C}
        -
        \|\boldsymbol{\mu}_C\|_2^2
    }.
\]

Since $\mathbf{S}_C$, $Q_C$, and $\mathbf{R}_C$ are additive under block
merging, both the purity criterion and pooled scoring statistics can be
updated using only block-level quantities.

\section{Experimental Setup and Additional Analysis}
\label{app:additional_analysis}

\subsection{Evaluation Setup}
\label{app:setup}
\paragraph{Benchmarks.} We evaluate HASA on two complementary long-context benchmarks: \textbf{LongBench-Pro} \citep{longbench} and \textbf{RULER-32K} \citep{ruler}. LongBench-Pro contains 1,500 English and Chinese examples spanning 11 primary and 25 secondary tasks, with input lengths extending to 256K tokens. It evaluates long-context understanding across different languages and difficulty levels. RULER-32K provides a controlled evaluation at a 32K context length, covering retrieval, multi-hop tracing, aggregation, and question-answering tasks. Together, the two benchmarks test both realistic long-context reasoning and controlled retrieval over extended contexts.

\paragraph{Baselines.} We compare HASA against four representative training-free sparse-attention methods. \textbf{H$_2$O} retains tokens according to accumulated historical attention importance; \textbf{Quest} performs query-aware retrieval over fixed KV pages using per-page min/max key statistics; \textbf{InfLLM} retrieves relevant context units from a long-term memory representation; and \textbf{UNIQUE}, the closest baseline to our setting, represents each fixed KV page using its mean key vector and scalar dispersion for query-dependent selection. We also report \textbf{dense attention} as the full-context reference. All sparse
methods are evaluated without additional model training or fine-tuning.

\paragraph{Evaluation Protocol.}
Experiments use Ministral-3-8B-Instruct-2512 for the main evaluation, with Qwen3-8B used as an additional backbone in Appendix~\ref{app:additional_eval}. This allows us to assess whether the proposed selection mechanism generalizes across different model architectures. We report the individual task scores and their average. Sparse methods are compared under matched per-query KV budgets of \textbf{512}, \textbf{1024}, and \textbf{2048} tokens during decoding, while dense attention accesses the full available KV context. We follow the corresponding benchmark evaluation protocols and use the same model, input, decoding, and KV-budget settings across sparse methods to ensure a matched comparison. Unless otherwise specified, HASA uses a base region size of $S_0=8$, $\lambda=0.5$, $\epsilon_{\max}=0.10$, $\rho_{\min}=0.70$, and $C_{\max}=512$.

\subsection{Additional Backbone Evaluation}
\label{app:additional_eval}

\begin{table}[t]
\centering
\caption{Accuracy results on LongBench-Pro using Qwen3-8B. The KV budget denotes the number of KV tokens available to each query during decoding; full attention is unrestricted.}
\label{tab:longbench_qwen}
\resizebox{0.9\columnwidth}{!}{
\begin{tabular}{l|c|c|cc|cccc}
\toprule
\textbf{Method} & \textbf{Budget} &
\textbf{Overall} & \textbf{En} & \textbf{Zh} &
\textbf{Extreme} & \textbf{Hard} &
\textbf{Moderate} & \textbf{Easy} \\
\midrule

Full attention & -- &
37.13 & 35.92 & 38.34 & 27.49 & 35.21 & 31.44 & 46.86 \\

\midrule

%
Quest & 2048 &
35.02 & 33.49 & 36.55	& 25.62	& 33.29	& 29.33	& 44.51 \\

UNIQUE & 2048 &
36.12 & 35.03 & 37.22 & 27.15 & 33.49 & 29.85 & 45.99 \\

HASA (Ours) & 2048 &
\textbf{36.64} & \textbf{35.77} & \textbf{37.50} &
\textbf{27.62} & \textbf{34.32} & \textbf{30.06} & \textbf{46.53} \\

\midrule

Quest & 1024 &
34.31 & 33.14 & 35.47 & 24.71 & \textbf{33.24} & 27.94 & 43.92 \\

UNIQUE & 1024 &
35.16 & \textbf{34.17} & 36.15 & \textbf{26.62} & 32.46 & 28.41 & 44.99 \\

HASA (Ours) & 1024 &
\textbf{35.63} & 33.85 & \textbf{37.42} & 26.27 & 33.02 & \textbf{29.73} & \textbf{45.59} \\

\midrule

Quest & 512 &
32.87 & 31.36 & 34.37 & 23.69 & 28.77 & 25.24 & 44.13 \\

UNIQUE & 512 &
34.49 & \textbf{33.41} & 35.56 & 24.88 & \textbf{33.37} & 27.74 & \textbf{44.31} \\

HASA (Ours) & 512 &
\textbf{34.84} & 32.41 & \textbf{37.27} & \textbf{26.60} & 32.15 & \textbf{28.74} & 44.17 \\

\bottomrule
\end{tabular}
}
\end{table}

\begin{table*}[t]
\centering
\caption{Accuracy results on RULER-32K using Qwen3-8B. The KV budget denotes the number of KV tokens available to each query during decoding; full attention is unrestricted.}
\label{tab:ruler_qwen}
\resizebox{\textwidth}{!}{
\begin{tabular}{l|c|ccccccccccccc|c}
\toprule
\textbf{Method} & \textbf{Budget} &
\textbf{SG1} & \textbf{SG2} & \textbf{SG3} & \textbf{MK1} &
\textbf{MK2} & \textbf{MK3} & \textbf{MV} &
\textbf{MQ} & \textbf{VT} & \textbf{CWE} &
\textbf{FWE} & \textbf{QA1} & \textbf{QA2} &
\textbf{Avg.} \\
\midrule

Dense & -- &
100.00 & 100.00 & 100.00 & 100.00 & 97.00 & 95.00 
& 99.75 & 99.75 & 99.80 & 84.10 & 94.67 & 75.00 & 50.00 & 91.93 \\

\midrule

Quest & 2048 &
\textbf{100.00} & \textbf{100.00} & \textbf{100.00} & 99.00 & \textbf{98.00} & 78.00 
& \textbf{100.00} & 99.00 & 98.20 & 58.60 & 90.67 & \textbf{75.00} & 55.00 & 88.57 \\

UNIQUE & 2048 &
\textbf{100.00} & \textbf{100.00} & \textbf{100.00} & 99.00 & \textbf{98.00} & 79.00 
& \textbf{100.00} & 99.25 & 98.20 & 59.50 & 90.67 & \textbf{75.00} & \textbf{58.00} & 88.97 \\

HASA (Ours) & 2048 &
\textbf{100.00} & \textbf{100.00} & \textbf{100.00} & \textbf{100.00} & 97.00 & \textbf{84.00} 
& 99.75 & \textbf{99.75} & \textbf{99.80} & \textbf{71.40} & \textbf{93.67} & 74.00 & 56.00 & \textbf{90.41} \\

\midrule

Quest & 1024 &
\textbf{100.00} & \textbf{100.00} & \textbf{100.00} & \textbf{100.00} & 96.00 & 44.00 
& 99.75 & 99.25 & \textbf{100.00} & \textbf{61.00} & \textbf{93.67} & 71.00 & 53.00 & 85.97 \\

UNIQUE & 1024 &
\textbf{100.00} & \textbf{100.00} & \textbf{100.00} & 99.00 & \textbf{97.00} & 62.00 
& 98.75 & \textbf{99.75} & 96.80 & 48.10 & 88.67 & 72.00 & \textbf{56.00} & 86.01 \\

HASA (Ours) & 1024 &
\textbf{100.00} & \textbf{100.00} & \textbf{100.00} & 97.00 & 88.00 & \textbf{82.00} 
& \textbf{100.00} & 96.75 & 96.80 & 56.10 & 88.67 & \textbf{73.00} & 50.00 & \textbf{86.79} \\

\midrule

Quest & 512 &
\textbf{100.00} & \textbf{100.00} & \textbf{100.00} & \textbf{100.00} & 90.00 & 7.00 
& 96.50 & 97.75 & \textbf{99.60} & \textbf{44.90} & \textbf{95.33} & 70.00 & 49.00 & 80.78 \\

UNIQUE & 512 &
\textbf{100.00} & \textbf{100.00} & 98.00 & 99.00 & \textbf{97.00} & 41.00 
& 97.75 & \textbf{99.00} & 95.60 & 35.30 & 87.67 & 73.00 & \textbf{55.00} & 82.95 \\

HASA (Ours) & 512 &
\textbf{100.00} & \textbf{100.00} & 99.00 & 96.00 & 81.00 & \textbf{66.00} 
& \textbf{98.75} & \textbf{99.00} & 95.20 & 43.20 & 81.33 & \textbf{74.00} & 52.00 & \textbf{83.50} \\

\bottomrule
\end{tabular}
}
\end{table*}

To further evaluate the effectiveness of HASA across model backbones, we conduct additional experiments on Qwen3-8B using the same evaluation protocol and matched KV budgets. As shown in Tables~\ref{tab:longbench_qwen} and~\ref{tab:ruler_qwen}, HASA achieves the highest overall scores among the evaluated sparse methods at every KV budget on both LongBench-Pro and RULER-32K.


On LongBench-Pro, HASA achieves overall scores of \textbf{36.64},
\textbf{35.63}, and \textbf{34.84} at KV budgets of 2048, 1024, and 512,
respectively, compared with 36.12, 35.16, and 34.49 for UNIQUE. The gains
therefore persist as the available KV budget is reduced, with improvements of 0.52, 0.47, and 0.35 points over the strongest sparse baseline. The language- and difficulty-level breakdowns show that the advantage is not restricted to a single subset: HASA improves substantially on Chinese at the lower budgets and remains competitive across the remaining categories.  A similar pattern is observed on RULER-32K. HASA obtains average scores of \textbf{90.41}, \textbf{86.79}, and \textbf{83.50} at budgets of 2048, 1024, and 512, respectively, outperforming UNIQUE by 1.44, 0.78, and 0.55 points. The improvement is particularly visible on several difficult multi-key and retrieval-sensitive tasks, while performance on already saturated retrieval tasks remains comparable across methods. Importantly, HASA continues to retain the highest overall score as the KV budget becomes more restrictive.

The consistent improvements across model backbones, benchmarks, and KV budgets demonstrate the broad effectiveness and generalizability of HASA. By deriving feature ranks and sequence granularity from each head's pretrained query--key interaction structure, HASA adapts its sparse selection to different models rather than relying on a fixed resolution configuration.


\subsection{Hyperparameter Analysis}
\label{app:hyperparameter}

\begin{table}[!tb]
\centering
\caption{Hyperparameter sensitivity on RULER-32K at a 1024-token KV budget. Each hyperparameter is varied independently while the remaining parameters are fixed to their best-performing values:
$\epsilon_{\max}=0.10$, $\rho_{\min}=0.70$, and $C_{\max}=512$.}
\label{tab:hyperparam_sensitivity}
\setlength{\tabcolsep}{8pt}
\begin{tabular}{lcc}
\toprule
\textbf{Hyperparameter} & \textbf{Value} & \textbf{Accuracy (\%)} \\
\midrule

\multirow{4}{*}{$\epsilon_{\max}$}
& 0.05 & 92.21 \\
& \textbf{0.10} & \textbf{92.21} \\
& 0.15 & 91.98 \\
& 0.20 & 91.04 \\
\midrule

\multirow{4}{*}{$\rho_{\min}$}
& 0.50 & 90.82 \\
& 0.60 & 91.73 \\
& \textbf{0.70} & \textbf{92.21} \\
& 0.80 & 92.16 \\
\midrule

\multirow{4}{*}{$C_{\max}$}
& 128  & 92.16 \\
& 256  & 92.17 \\
& \textbf{512}  & \textbf{92.21} \\
& 1024 & 91.42 \\
\bottomrule
\end{tabular}
\end{table}

Table~\ref{tab:hyperparam_sensitivity} examines the sensitivity of HASA to the three hyperparameters controlling feature and sequence resolution. We use RULER-32K and the intermediate 1024-token KV budget as a representative setting, varying each hyperparameter independently while holding the others fixed. Overall, performance remains stable over a reasonably broad range around the default configuration.

For the spectral residual threshold $\epsilon_{\max}$, HASA achieves the same reported accuracy (92.21\%) at 0.05 and 0.10. We adopt 0.10 as the default because its more permissive residual constraint can reduce the retained feature rank and selection overhead without sacrificing accuracy in this setting. Increasing the threshold to 0.15 causes only a slight decline to 91.98\%, whereas a larger threshold of 0.20 reduces accuracy to 91.04\%. This indicates that excessively aggressive spectral truncation can discard interaction directions that remain useful for sparse selection.

The sequence-resolution parameters show a similar trade-off. Lowering the purity threshold $\rho_{\min}$ permits more aggressive aggregation of neighboring regions and reduces accuracy from 92.21\% at $\rho_{\min}=0.70$ to 90.82\% at 0.50. Increasing the threshold to 0.80 produces nearly identical performance (92.16\%), suggesting that $\rho_{\min}=0.70$ already preserves sufficient sequence resolution for reliable selection in this setting. Performance is likewise stable for $C_{\max}$ between 128 and 512 tokens, with scores ranging from 92.16\% to 92.21\%. Allowing substantially larger regions ($C_{\max}=1024$) lowers accuracy to 91.42\%, indicating that excessive aggregation can make region summaries too coarse for reliable sparse selection.

Together, these results support the intended behavior of the two adaptive mechanisms: overly aggressive feature truncation or sequence aggregation can remove useful selection detail, while HASA remains robust within a broad intermediate operating range.

\end{document}